\documentclass[pdflatex,sn-mathphys-num]{sn-jnl}% Math and Physical Sciences Numbered Reference Style
\usepackage{graphicx}%
\usepackage{multirow}%
\usepackage{amsmath,amssymb,amsfonts}%
\usepackage{amsthm}%
\usepackage{mathrsfs}%
\usepackage[title]{appendix}%
\usepackage{xcolor}%
\usepackage{textcomp}%
\usepackage{manyfoot}%
\usepackage{booktabs}%
\usepackage{algorithm}%
\usepackage{algorithmicx}%
\usepackage{algpseudocode}%
\usepackage{listings}%
\usepackage{subcaption}
\usepackage{tikz}

\theoremstyle{thmstyleone}%

\theoremstyle{thmstyletwo}%

\theoremstyle{thmstylethree}%

\begin{document}

% \title[Diffusion-Based Sub-Visible Particle Image Generation to Mitigate Data Imbalance for Machine Learning Approaches]{Diffusion-Based Sub-Visible Particle Image Generation to Mitigate Data Imbalance for Machine Learning Approaches}

\title[Improved Sub-Visible Particle Classification in Flow Imaging Microscopy via Generative AI-Based Image Synthesis]{Improved Sub-Visible Particle Classification in Flow Imaging Microscopy via Generative AI-Based Image Synthesis}

% Mitigating Data Imbalance and Improving Classification of Sub-Visible Particles Using Generative AI

% Generative AI-Driven Sub-Visible Particle Image Synthesis for Improved Particle Classification

% Generative AI-Based Image Synthesis for Improved Sub-Visible Particle Classification

\author*[1,2]{\fnm{Utku} \sur{Ozbulak}}\email{utku.ozbulak@ghent.ac.kr}

\author[3]{\fnm{Michaela} \sur{Cohrs}}

\author[4]{\fnm{Hristo L.} \sur{Svilenov}}

\author[1,5]{\fnm{Joris} \sur{Vankerschaver}}

\author[1,2]{\fnm{Wesley} \sur{De Neve}}

\affil[1]{\orgdiv{Center for Biosystems and Biotech Data Science}, \orgname{Ghent University Global Campus}, \orgaddress{\city{Incheon}, \country{Republic of Korea}}}

\affil[2]{\orgdiv{Department of Electronics and Information Systems}, \orgname{Ghent University}, \orgaddress{\city{Ghent}, \country{Belgium}}}

\affil[3]{\orgdiv{Faculty of Pharmaceutical Sciences}, \orgname{Ghent University}, \orgaddress{\city{Ghent}, \country{Belgium}}}

\affil[4]{\orgdiv{Biopharmaceutical Technology, TUM School of Life Sciences}, \orgname{Technical University of Munich}, \city{Freising}, \country{Germany}}

\affil[5]{\orgdiv{Department of Mathematics, Computer Science and Statistics}, \orgname{Ghent University}, \orgaddress{\city{Ghent}, \country{Belgium}}}

\abstract{
\let\thefootnote\relax\footnotetext{Preprint.}
Sub-visible particle analysis using flow imaging microscopy combined with deep learning has proven effective in identifying particle types, enabling the distinction of harmless components such as silicone oil from protein particles. However, the scarcity of available data and severe imbalance between particle types within datasets remain substantial hurdles when applying multi-class classifiers to such problems, often forcing researchers to rely on less effective methods. The aforementioned issue is particularly challenging for particle types that appear unintentionally and in lower numbers, such as silicone oil and air bubbles, as opposed to protein particles, where obtaining large numbers of images through controlled settings is comparatively straightforward. In this work, we develop a state-of-the-art diffusion model to address data imbalance by generating high-fidelity images that can augment training datasets, enabling the effective training of multi-class deep neural networks. We validate this approach by demonstrating that the generated samples closely resemble real particle images in terms of visual quality and structure. To assess the effectiveness of using diffusion-generated images in training datasets, we conduct large-scale experiments on a validation dataset comprising 500,000 protein particle images and demonstrate that this approach improves classification performance with no negligible downside. Finally, to promote open research and reproducibility, we publicly release both our diffusion models and the trained multi-class deep neural network classifiers, along with a straightforward interface for easy integration into future studies, at \href{https://github.com/utkuozbulak/svp-generative-ai}{github.com/utkuozbulak/svp-generative-ai}.
}


\maketitle

\section{Introduction}
Sub-visible particles (SvPs) are a critical quality concern in protein-based therapeutics due to their association with a range of adverse effects, including immunogenic responses, reduced drug efficacy, and compromised product stability~\cite{carpenter2009overlooking,chisholm2017immunogenicity}. The presence of particulate matter can trigger immune activation, potentially leading to anti-drug antibody formation or even anaphylaxis in severe cases~\cite{kotarek2016subvisible,uchino2017immunogenicity}.

Due to these safety and efficacy concerns, regulatory agencies have issued stringent guidelines to monitor and limit particulate matter in parenteral drugs. The United States Pharmacopeia (USP) and the European Pharmacopoeia mandate the quantification of particles $\geq 10$ µm and $\geq 25$ µm~\cite{USP787_2021,USP788_2025,EurPharm_2_9_19_2021}. However, these standards primarily address particle count, often overlooking the chemical nature or morphological characteristics of the particles, which are equally vital for assessing risk and informing root-cause analysis.

Sub-visible particles in biotherapeutics can arise from various sources and are commonly classified into three categories: inherent (consisting of product itself, e.g., protein aggregates), intrinsic (container or delivery system, e.g., glass or rubber), and extrinsic (external contaminants, e.g., fibers or dust)~\cite{allen2008dosage}. Certain types of SvPs have been directly linked to reduced therapeutic potency and long-term instability. For example, protein aggregates can cause immunogenic reactions and loss of activity. Meanwhile, other particles, such as particles consisting of silicone oil, which is often used as a lubricant in syringes, are generally considered as low toxic~\cite{melo2019release,krayukhina2019assessment}. Nevertheless, even particles that are considered non-toxic can act as catalysts that promote further protein aggregation, indirectly compromising product safety~\cite{zaman2014nanoparticles}. As a result, accurate detection and classification of SvPs has emerged as a key challenge in pharmaceutical quality control.

\begin{figure}[t!]
\centering
\includegraphics[width=1\textwidth]{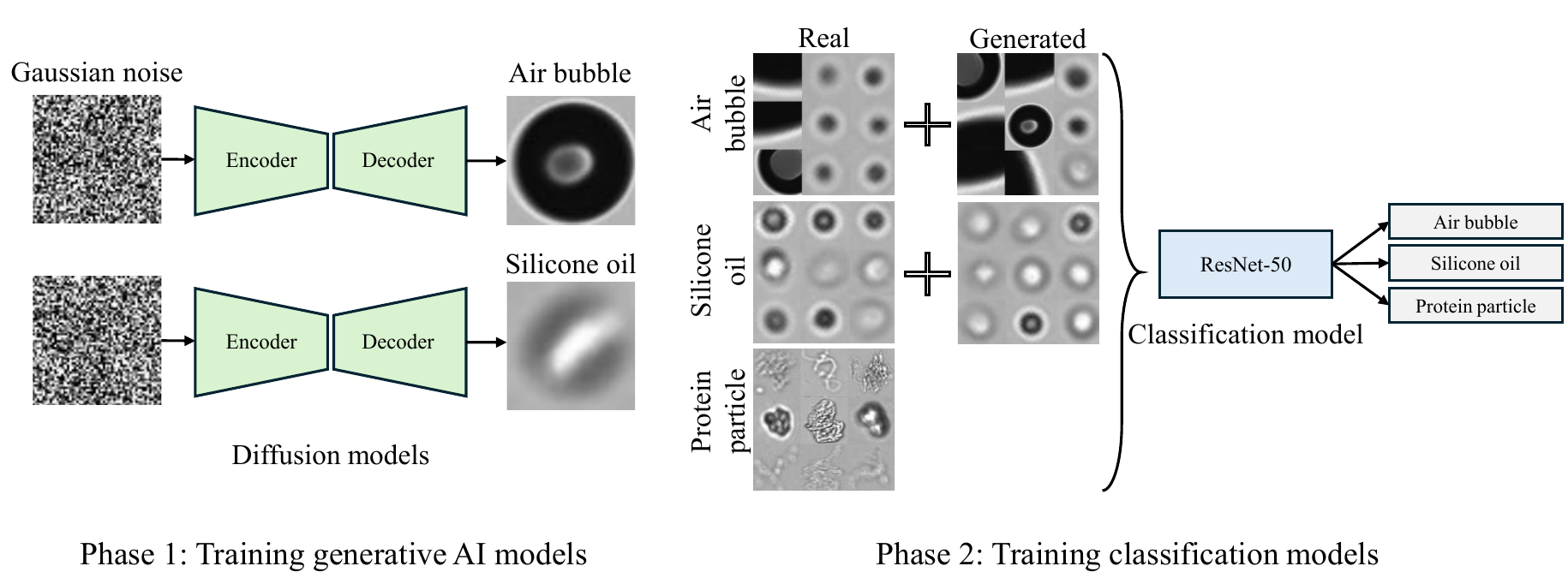}
\caption{Visual abstract for the main contribution of the paper. In Phase 1, we train a diffusion-based generative AI model to synthesize SvP images of underrepresented classes (silicone oil and air bubbles). In Phase 2, we train a multi-class classifier using the augmented dataset to demonstrate the usefulness of the generated images in improving classification performance.}
\label{fig:visual abstract}
\end{figure}

To address the challenge of identifying and characterizing SvPs, flow-imaging microscopy (FIM) in combination with machine learning methods -- particularly deep learning-based approaches -- has emerged as powerful tool~\cite{cohrs2024color,grabarek2021particulate,lopez2024sub,gambe2020automatic}. These machine learning techniques can learn complex visual patterns, including morphological ones directly from imaging data, enabling automated classification of particle types with greater accuracy compared to traditional rule-based or manual methods~\cite{VGG,calderon2017using}. However, one of the main obstacles in applying deep learning to FIM images is the need for large datasets with balanced numbers of images for the different classes~\cite{nakae2025application}. Deep learning models are notoriously brittle when trained on limited data and are highly sensitive to class imbalance~\cite{pecher2024survey}. To address this, some research efforts have explored one-class classifiers, which are more robust under imbalanced conditions, although multi-class classifiers typically offer higher overall classification accuracy~\cite{nakae2025application,lopez2024sub}.

Data scarcity for SvPs is primarily an issue for certain particle types, such as silicone oil droplets and air bubbles, since accurate classification of these particles often requires extensive manual labeling by subject-matter experts~\cite{lopez2024sub}. In contrast, acquiring large datasets for protein aggregates is relatively straightforward, as particles can be generated under stress. This creates a unique challenge: despite the availability of abundant data for certain particle types, the overall utility of these datasets is limited in machine learning approaches due to the severe imbalance caused by underrepresented particle types. As a result, even large datasets may fail to support robust multi-class learning, highlighting the need for methods that can overcome this hurdle, or generate and augment data for minority classes to ensure balanced and effective model training.

In this work, we address the issue of data imbalance within subvisible particle image datasets by developing a diffusion-based generative model~\cite{denoising_diffusion} inspired by recent state-of-the-art architectures that have become foundational to many generative image products, including Stability AI's Stable Diffusion~\cite{stable_diffusion}, OpenAI's Sora~\cite{sora}, and Google Gemini~\cite{gemini}. Our focus is on generating high-fidelity images of underrepresented particle types, particularly silicone oil droplets and air bubbles, which are challenging to annotate and acquire at scale. By synthetically generating realistic images for these minority classes, our approach enables the full utilization of large SvP datasets without introducing imbalance-related biases, thereby supporting more robust and accurate multi-class classification. An overview of the two-phase approach is presented in Figure~\ref{fig:visual abstract}: in the first phase, we train a diffusion model to generate high-fidelity synthetic images, and in the second phase, we use these generated images to supplement the training of a multi-class classifier. To further support the community and promote open science, we publicly release our diffusion model, along with sample datasets and implementation code to facilitate its adoption in pharmaceutical quality control and related research applications.

\section{Materials and Methods}

\subsection{Materials}\label{sec:materials}
Eight commercial monoclonal antibodies (mAbs) were used to generate protein aggregates. mAbs were formulated in 10 mM acetate buffer, pH 5 at 0.5 mg/ml for heat stress or at 1 mg/ml in 10 mM histidine buffer, pH 6, with 0.9\% NaCl and 0.05\% polysorbate 20 for mechanical stress. 
0.9\% NaCl was used to recover silicone oil droplets from siliconized syringes (5 ml Plastipak Luer Lock, BD, Franklin Lakes, US). All solutions were filtered through 0.22 µm filters before use. All chemicals were pharma grade or higher.

\subsection{Sample Preparation and Measurement}\label{sec:sample}

Protein aggregates were produced from different mAb solutions via heat or mechanical stress to achieve various morphologies. A digital heat block (VWR, Radnor, USA) was used to heat mAbs in 2 ml microcentrifuge tubes at 90 ° C for 5 or 20 min. Mechanical stress was produced via orbital shaking of 1.35 ml mAb in 2R precleaned glass vials at 300 rpm with 19 mm orbit on a digital orbital shaker (Heathrow Scientific, Illionois, USA) for 48 h.

Silicone oil droplets were recovered from syringes by filling them with 5 ml of 0.9\% NaCl and subjecting them to thourough manual agitation for 2 min. 200 µl of each sample were imaged in triplicates using flow-imaging microscopy (FlowCam 8000, Yokogawa Fluid Imaging Technologies, Scarborough USA). The instrument was equipped with a 10x objective and a 80 x 700 µm flow cell. A flowrate of 0.15 ml/s in combination with an auto image frame rate of 27 frames/s was used. Particles were captured when 13/10 was exceeded for dark and light pixels. Particles were separated as of 3 µm distance to nearest neighbor. 

Images of air bubbles were manually identified within all samples and separated from the dataset. Those air bubbles were used for machine learning while the remaining dataset was labeled as protein aggregates and silicone oil and also used as described below.

\subsection{Dataset Preparation}\label{sec:dataset}

Using the aforementioned sample preparation protocols, we acquired a total of 520,000 protein particle images as well as 1,500 images of silicone oil and 1,500 images of air bubbles. Due to the resolution variability of the FIM images, we first standardized all images by resizing their smallest edge to 64 pixels. To do that, we followed the normalization procedure described in~\cite{nakae2025application}. This ensures a minimum image size of $64 \times 64$ while preserving the aspect ratio. % Figure~\ref{fig:image_cnt} shows the distribution of particle image dimensions prior to resizing, with each point representing an individual image in terms of its width and height.

\textbf{Training Data for Diffusion Models.}
To prevent data leakage between training and validation, diffusion models are trained exclusively on a separate set of 1,000 real images for each minority class (silicone oil and air bubbles), separate from those used in evaluation. These images served as the basis for learning the generative distribution of underrepresented SvP types.

\textbf{Validation Data for Classification Models.}
To provide a consistent and unbiased evaluation baseline, we allocated a fixed validation set comprising 500,000 protein particle images and 500 images each for silicone oil and air bubbles. This heavily imbalanced setup (1000:1 ratio between protein and other classes) reflects real-world class skew and allowed us to robustly assess the predictive performance of models under challenging conditions.

\textbf{Training Data for Classification Models.}
From the remaining data (20,000 protein images and 1000 images of silicone oil and air bubbles), as well as the generated data using diffusion models, we constructed multiple training configurations to evaluate the effect of dataset size on classification performance. These configurations fall into two categories below and are summarized in Table~\ref{tbl:dataset_summary}.

\begin{itemize}
    \item \textbf{Real-n}: Training datasets consisting of only real particle images. The number of samples per class increases progressively across Real-0 to Real-4, ranging from fewer images to a larger number of images with increased data imbalance.
    \item \textbf{Mixed-n}: Training datasets augmented with synthetic images generated via diffusion models. Each Mixed-n set mirrors its Real-n counterpart in real sample count but supplements minority classes (air bubbles and silicone oil) with corresponding generated samples, balancing class distributions while enhancing data diversity.
\end{itemize}

\begin{table}[t]
\centering
\caption{Summary of training datasets used for the classification models in this study. Each row represents a different dataset configuration. ``Mixed" datasets contain both real and AI-generated images, while ``Real" datasets include only real FIM images. All counts are reported in thousands (K), where 1K = 1,000 images.}
\scriptsize
\label{tbl:dataset_summary}
\begin{tabular}{ccc|cc|c|c}
\toprule
\multirow{2}{*}{\shortstack{Dataset\\Descriptor}} 
 & \multicolumn{2}{c}{Silicone Oil} 
 & \multicolumn{2}{c}{Air Bubble} 
 & Protein Particle 
 & \multirow{2}{*}{\shortstack{Class\\Balance}} \\
\cmidrule[0.75pt]{2-6}
~ & Real & Generated & Real & Generated & Real & \\
\midrule
Real-0 & 1K & 0 & 1K & 0 & 1K & Balanced \\
\midrule
Real-1 & \multirow{4}{*}{\shortstack{1K}} & \multirow{4}{*}{\shortstack{0}} & \multirow{4}{*}{\shortstack{1K}} & \multirow{4}{*}{\shortstack{0}} & 2K & \multirow{4}{*}{Imbalanced} \\
Real-2 & ~ & ~ & ~ & ~ & 5K & ~ \\
Real-3 & ~ & ~ & ~ & ~ & 10K & ~ \\
Real-4 & ~ & ~ & ~ & ~ & 20K & ~ \\
\midrule
Mixed-1 & \multirow{4}{*}{\shortstack{1K}} & 1K & \multirow{4}{*}{\shortstack{1K}} & 1K & 2K & \multirow{4}{*}{Balanced} \\
Mixed-2 & ~ & 4K & ~ & 4K & 5K & ~ \\
Mixed-3 & ~ & 9K & ~ & 9K & 10K & ~ \\
Mixed-4 & ~ & 19K & ~ & 19K & 20K & ~ \\
\bottomrule
\end{tabular}
\end{table}

\subsection{Diffusion Models}\label{sec:diffusion_models}

Generative diffusion models are a class of machine learning models designed to generate new data (such as images, sound, or text) by progressively transforming random noise into a structured, meaningful output~\cite{diffusion_working,denoising_diffusion}. These models are based on the concept of diffusion, which refers to the gradual mixing or spreading of particles in a medium~\cite{diffusion_early}. Generative diffusion models consist of two main steps: the forward process (diffusion), where noise is progressively added to the data over time, and the reverse process (denoising), where the model learns to reverse this noise addition, gradually recovering the original data. The forward process transforms the data into pure noise, and the reverse process generates new, meaningful data by denoising the corrupted input, ultimately producing realistic samples.

\textbf{Forward Process}. In the forward process, noise is progressively added to the data (e.g., an image) over several steps, eventually converting it into pure noise. This process, which does not involve a neural network, adds structured noise, such as Gaussian noise. In Denoising Diffusion Probabilistic Models (DDPMs), the type of models we use in this work, the forward process is a Markov chain where noise is added at each step, and the reverse process is learned as a probabilistic model to denoise the data step by step~\cite{denoising_diffusion}.

Let \( \mathbf{x}_0 \) represent the original data to which at each step \( t \) Gaussian noise is added. The result after \( t \) steps, \( \mathbf{x}_t \), is given by:

\[
\mathbf{x}_t = \sqrt{\alpha_t} \mathbf{x}_{t-1} + \sqrt{1 - \alpha_t} \mathbf{\epsilon}_t
\]

In this equation, \( \mathbf{x}_t \) is the noisy data at step \( t \), \( \alpha_t \) is a hyperparameter controlling the strength of the noise added at each step, typically decaying over time (i.e., \( \alpha_t \to 0 \) as \( t \) increases), and \( \mathbf{\epsilon}_t \) is Gaussian noise sampled from a normal distribution, \( \mathbf{\epsilon}_t \sim \mathcal{N}(0, I) \). The forward diffusion process progressively corrupts the data, and as \( t \to T \), the data becomes increasingly noisy, ultimately approaching pure random noise.

\begin{figure}[t!]
\centering
\begin{subfigure}{0.45\textwidth}
\includegraphics[width=1\textwidth]{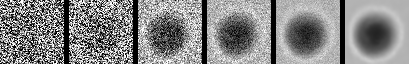}
\includegraphics[width=1\textwidth]{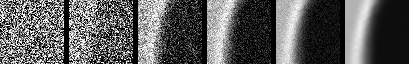}
\includegraphics[width=1\textwidth]{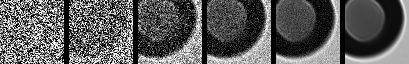}
\includegraphics[width=1\textwidth]{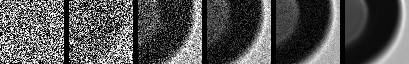}
\caption{Air bubbles}
\end{subfigure}
\begin{subfigure}{0.45\textwidth}
\includegraphics[width=1\textwidth]{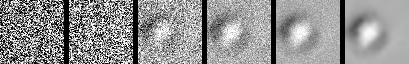}
\includegraphics[width=1\textwidth]{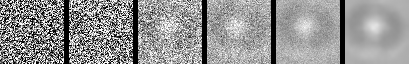}
\includegraphics[width=1\textwidth]{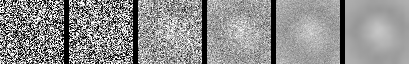}
\includegraphics[width=1\textwidth]{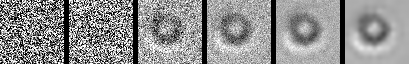}
\caption{Silicone oil droplets}
\end{subfigure}
\caption{Visualization of the reverse diffusion process for (a) air bubbles and (b) silicone oil droplets. Each row depicts selected time steps showing the progressive denoising of random noise into structured and realistic sub-visible particle images.}
\label{fig:image_diffusion_process}
\end{figure}

\textbf{Reverse Process}. The reverse diffusion process is the core of the generative model. It aims to reverse the noising process by learning the probability distribution \( p_{\theta}(\mathbf{x}_{t-1} \mid \mathbf{x}_t) \) for each step \( t \), which is given by:

\begin{equation}
p_{\theta}(\mathbf{x}_{t-1} \mid \mathbf{x}_t) = \mathcal{N}(\mathbf{x}_{t-1}; \mu_{\theta}(\mathbf{x}_t, t), \sigma_t^2 I) \,. 
\end{equation}

Here, \( \mu_{\theta}(\mathbf{x}_t, t) \) is the predicted mean of the model for the denoised data at step \( t-1 \). The goal is to learn the parameters \( \theta \) that allow the model to recover clean data from noisy inputs.

\textbf{Image Generation}. The image generation process begins with a sample of pure random Gaussian noise, denoted as \( \mathbf{x}_T \). The goal is to progressively reverse the noising process, step by step, using the learned model to generate a realistic image. As shown in Figure~\ref{fig:image_diffusion_process}, the model applies the learned denoising process starting from the random noise, iteratively refining the noisy image at each step to reconstruct the original data distribution.

At each step \( t \), the model predicts the denoised data \( \mathbf{x}_{t-1} \) from the noisy data \( \mathbf{x}_t \) by sampling from the learned distribution \( p_{\theta}(\mathbf{x}_{t-1} \mid \mathbf{x}_t) \), as defined in the reverse process. This process continues until \( t = 0 \), at which point the generated data \( \mathbf{x}_0 \) is a sample from the target distribution, resembling a clean image. The final result is an image that was generated from random noise, having gone through a series of denoising steps, guided by the reverse diffusion process learned during training.

\subsection{Model Evaluation}

We set up our diffusion model and training environment as described below.

\textbf{Diffusion Model}.
Following the work of~\cite{denoising_diffusion}, we used a U-Net architecture with several modifications~\cite{ronneberger2015u}. First, we replaced standard convolutional layers with weight-standardized convolutional layers, which have been shown to improve training stability and convergence~\cite{weight_standardize}. Second, we incorporated a combination of self-attention and linear attention layers, as these have been demonstrated to enhance the fidelity of generated images~\cite{attention,linear_attention}. Lastly, we applied Group Normalization before each attention layer~\cite{group_norm} to further stabilize training and improve feature normalization.

The forward process described in Section~\ref{sec:diffusion_models} consists of multiple noising time steps ($T$). In our setup, we found that using $T = 1,000$ steps with a linear noise schedule provides the best trade-off between image quality and computational efficiency.

\textbf{Diffusion Evaluation}.
To train the diffusion models, we use the $\text{L}_1$ loss between the predicted and ground truth noise at each timestep, which not only encourages sharper reconstructions but also improves the preservation of fine-grained structures in SvP images, which is crucial for distinguishing subtle morphological differences.

The $\text{L}_1$ loss is defined as follows: $\mathcal{L} = \left\| \hat{\epsilon} - \epsilon \right\|_1 = \sum_{i} \left| \hat{\epsilon}_i - \epsilon_i \right|$ where \( \hat{\epsilon} \) is the predicted noise, \( \epsilon \) is the ground truth noise, and the summation is over all pixel indices \( i \) in the image.

To quantitatively assess the realism of generated images throughout training, we additionally computed the Fréchet Inception Distance (FID). FID compares the distribution of generated images to that of real images in a deep feature space, providing a standard measure of generative quality~\cite{lucic2018gans,frechlet_dist,Seitzer2020FID}. With the setup described above, we trained our diffusion model for 1,000 epochs on the dataset detailed in Section~\ref{sec:dataset}. A full training cycle for a single model takes approximately 20 hours on a single NVIDIA A6000 GPU.

\begin{figure}[t!]
\centering
\begin{tikzpicture}
% Shift the matrix grid down to leave room for rotated labels
\begin{scope}[yshift=-0.5cm]
% Draw 3x3 grid
\draw (0,0) grid (3,3);

% Cell contents with C_{ij} notation
\node at (0.5,2.5) {\( C_{11} \)};
\node at (1.5,2.5) {\( C_{12} \)};
\node at (2.5,2.5) {\( C_{13} \)};

\node at (0.5,1.5) {\( C_{21} \)};
\node at (1.5,1.5) {\( C_{22} \)};
\node at (2.5,1.5) {\( C_{23} \)};

\node at (0.5,0.5) {\( C_{31} \)};
\node at (1.5,0.5) {\( C_{32} \)};
\node at (2.5,0.5) {\( C_{33} \)};

% Y-axis (true labels)
\foreach \i/\label in {0/Silicone oil,1/Air bubble,2/ Protein} {
\node[anchor=east] at (-0.2,2.5-\i) {\label};
}
\node[rotate=90] at (-2.5, 1.5) {[True class]};
\end{scope}

% X-axis (predicted labels), rotated
\foreach \i/\label in {0/Silicone oil,1/Air bubble,2/ Protein} {
\node[rotate=30, anchor=south] at (0.5+\i,3) {\label};
}
\node at (1.5, 4.0) {[Predicted class]};
\end{tikzpicture}

\caption{Confusion matrix illustrating the classification results for three particle types: protein particles, air bubbles, and silicone oil. Each entry \( C_{ij} \) represents the number of samples with true class \( i \) predicted as class \( j \). Diagonal entries \( C_{ii} \) indicate correct classifications, while off-diagonal entries represent misclassifications.}
\label{fig:confusion_matrix}
\end{figure}

\textbf{Classification Model}.
To evaluate the effectiveness of using diffusion-generated SVP images, we trained two multi-class classification models: a lightweight ResNet-18 and a larger ResNet-50~\cite{resnet}. These models are trained on the training splits described in Table~\ref{tbl:dataset_summary} and evaluated on the validation set detailed in Section~\ref{sec:dataset}. We applied a grid search strategy using both Adam and AdamW optimizers, exploring ten learning rates $\{10^{-5},\ 5 \times 10^{-4},\ 10^{-4},\ 5 \times 10^{-3},\ 10^{-3},\ 5 \times 10^{-2},\ 10^{-2}\}$ and three weight decay values $\{10^{-5}$, $10^{-4}$, $10^{-3}\}$, with batch sizes of 32, 64, or 128. This approach results in 126 training outcomes for each model and for each dataset split~\cite{adaptive_momentum,loshchilov2017decoupled}.

\textbf{Classification Evaluation}.
To assess classification performance, we report several evaluation metrics computed on an imbalanced validation set, where protein particles vastly outnumber silicone oil and air bubbles.

Given a confusion matrix in the form of Figure~\ref{fig:confusion_matrix}, let \( C_{ij} \) denote the entry of the confusion matrix corresponding to the number of samples whose \textbf{true class} is \( i \) and \textbf{predicted class} is \( j \). Correct classifications appear on the diagonal, i.e., \( C_{ii} \) represents the number of particles correctly predicted as class \( i \). Off-diagonal elements \( C_{ij} \) for \( i \neq j \) correspond to misclassifications from class \( i \) to class \( j \).

Given the severe class imbalance, we selected evaluation metrics that are robust to skewed distributions and provide a fair assessment across all classes, including underrepresented ones:

\begin{itemize}
\item \textbf{Class-based Precision}. Precision (Prec.) is calculated separately for each class to evaluate how many predicted particles of a given type are correctly identified.
\begin{equation}
\text{Precision}_i = \frac{C_{ii}}{\sum_{j=1}^3 C_{ji}} \quad \text{for } i = 1, 2, 3
\end{equation}

\item \textbf{Macro-averaged Class-based Precision}. This metric averages the class-wise precision scores equally across all classes, ensuring that minority classes like air bubbles and silicone oil are not overshadowed by the dominant protein class.
\begin{equation}
\text{Macro-Precision} = \frac{1}{3} \sum_{i=1}^{3} \text{Precision}_i
\end{equation}
\item \textbf{Precision-Recall Area Under Curve (AUPRC)}. This metric evaluates the trade-off between precision and recall across thresholds and is especially informative for rare classes, as it highlights how well the model distinguishes them from the majority class.

\begin{equation}
\text{AUPRC} = \int_0^1 \text{Precision}(r) \, dr
\end{equation}

where \( \text{Precision}(r) \) is the precision as a function of recall \( r \). The curve is typically constructed by varying the classification threshold and computing precision and recall at each point.
\end{itemize}

\begin{figure}[t!]
\centering
\begin{subfigure}{0.45\textwidth}
\includegraphics[width=\textwidth]{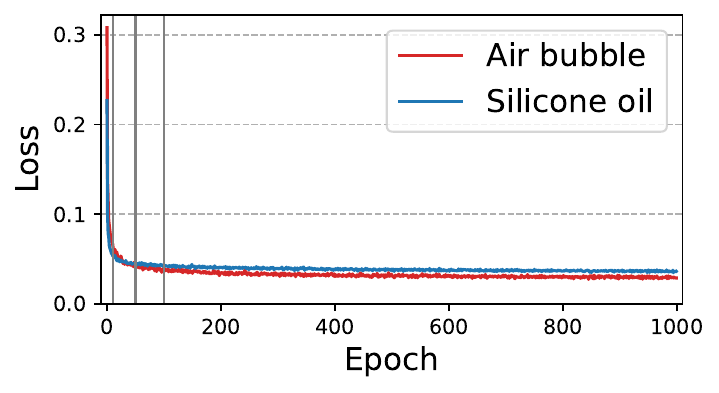}
\caption{Training loss}
\label{fig:loss_graph}
\end{subfigure}
\begin{subfigure}{0.45\textwidth}
\includegraphics[width=\textwidth]{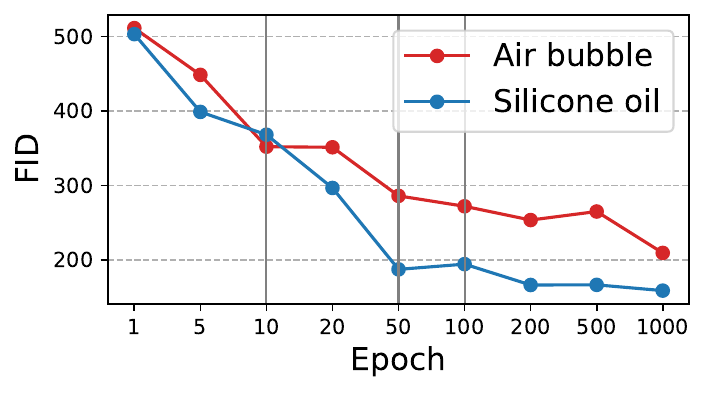}
\caption{FID}
\label{fig:fid_graph}
\end{subfigure}
\begin{subfigure}{0.45\textwidth}
\includegraphics[width=0.32\textwidth]{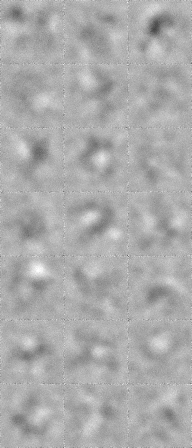}
\includegraphics[width=0.32\textwidth]{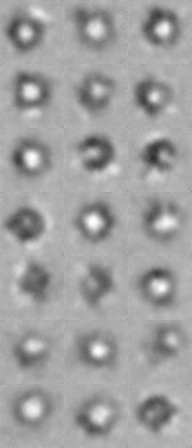}
\includegraphics[width=0.32\textwidth]{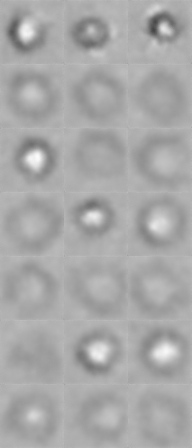}
\caption{Silicone oil droplets}
\label{fig:early_diffusion_silicone}
\end{subfigure}
\hspace{1em}
\begin{subfigure}{0.45\textwidth}
\includegraphics[width=0.32\textwidth]{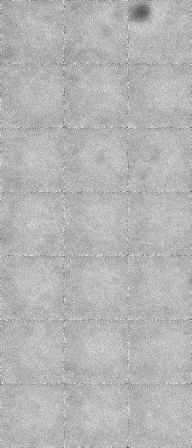}
\includegraphics[width=0.32\textwidth]{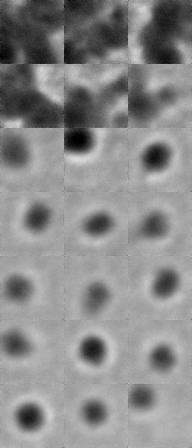}
\includegraphics[width=0.32\textwidth]{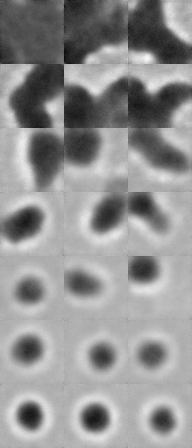}
\caption{Air bubbles}
\label{fig:early_diffusion_air}
\end{subfigure}
\caption{(a) Training loss curves for the diffusion models. At selected epochs, we generate sample images and compute the Fréchet Inception Distance (FID) between generated and real training images to quantitatively assess generative quality, as displayed in (b). Diffusion-generated images for (c) silicone oil droplets and (d) air bubbles, sampled at epochs 10, 50, and 100, showcasing low-fidelity images that do not resemble real FIM images. The corresponding sampling points are marked in (a) and (b) to contextualize image quality relative to training progression.}
\label{fig:diffusion_training}
\end{figure}

\section{Experimental Results}

\subsection{Diffusion Model Performance}

To evaluate the effectiveness of our generative diffusion model in synthesizing realistic SvP images, we first trained separate models for the two most underrepresented particle classes: air bubbles and silicone oil droplets. As described in Section~\ref{sec:dataset}, each model was trained using only 1,000 real images, carefully excluded from the validation set to avoid data leakage.

In \figurename~\ref{fig:loss_graph}, we present the training loss for the two diffusion models trained with silicone oil images and air bubble images. Alongside this, \figurename~\ref{fig:fid_graph} shows the FID scores computed from 100 generated images at various training epochs (1, 5, 10, 20, 50, 100, 200, 500, 1000). As can be seen, although the training loss plateaus around 0.2, the FID continues to improve over time, indicating that the perceptual quality of the generated samples keeps increasing even after the loss converges. This discrepancy suggests that traditional loss values may not fully reflect improvements in sample fidelity, whereas FID provides a more informative measure of generative quality, particularly in capturing subtle visual cues and morphological accuracy relevant to our application. It is important to note, however, that FID values are not inherently meaningful in isolation and should be interpreted relative to the domain: while natural image datasets typically yield lower FID scores, medical and biomedical images -- such as our SvP dataset -- are expected to produce higher scores~\cite{woodland2024feature,hashmi2024xreal}.

To visualize how this progression manifests in the output quality, we highlight three training epochs in \figurename~\ref{fig:loss_graph}, from which we sample and present generated images in \figurename~\ref{fig:early_diffusion_silicone} and \figurename~\ref{fig:early_diffusion_air}. As shown in these figures, early-stage outputs (e.g., epoch 5 or 10) are low-fidelity, often exhibiting incomplete particle shapes, indistinct boundaries, and missing structural features. In contrast, by epoch 50, the models generate more realistic and coherent particle images, capturing essential characteristics such as spherical morphology, texture, and transparency. This visual progression illustrates how the diffusion models incrementally refine their internal representation of sub-visible particle structure as training advances.

\begin{figure}[t!]
\centering
\begin{subfigure}{0.48\textwidth}
\includegraphics[width=\textwidth]{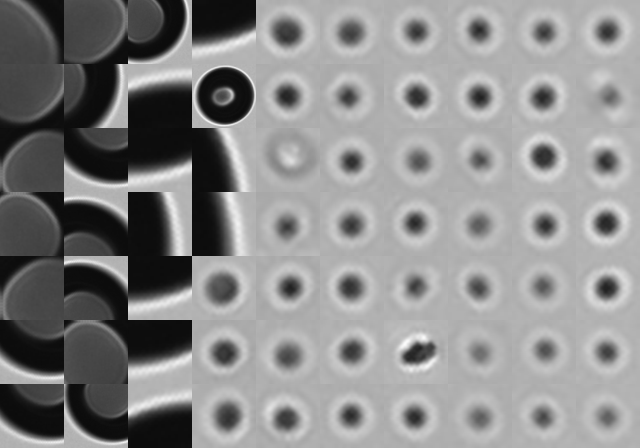}
\caption{Real air bubbles}
\end{subfigure}
\begin{subfigure}{0.48\textwidth}
\includegraphics[width=\textwidth]{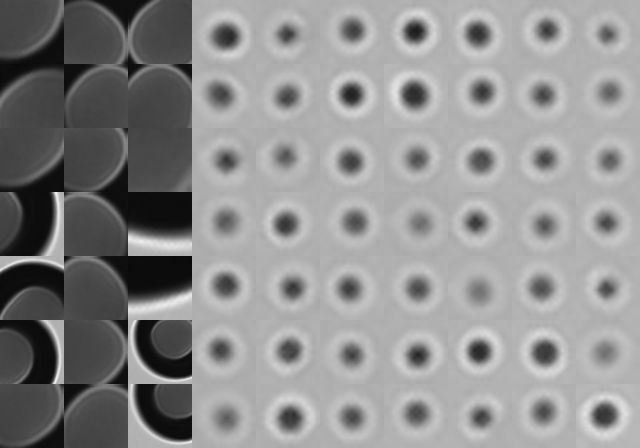}
\caption{Generated air bubbles}
\end{subfigure}
\begin{subfigure}{0.48\textwidth}
\includegraphics[width=\textwidth]{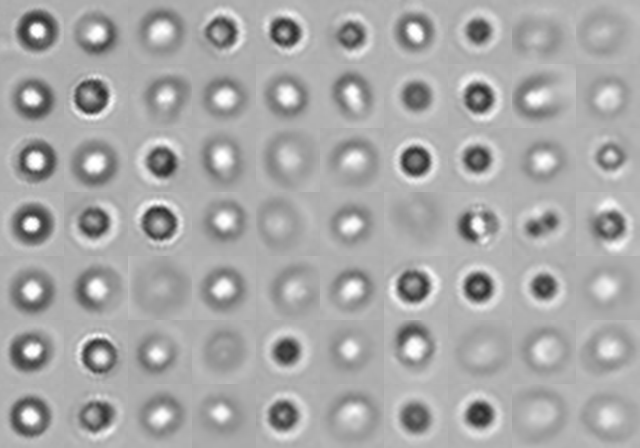}
\caption{Real silicone oil droplets}
\end{subfigure}
\begin{subfigure}{0.48\textwidth}
\includegraphics[width=\textwidth]{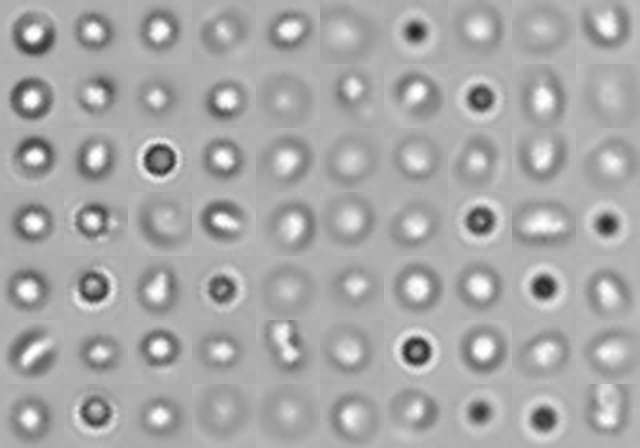}
\caption{Generated silicone oil droplets}
\end{subfigure}
\caption{Examples of diffusion-generated FIM images for underrepresented sub-visible particle types: (b) air bubbles and (d) silicone oil droplets compared to (a-c) real images obtained from FIM. Images on the right are generated by our diffusion models and the ones on the left are FIM images.}
\label{fig:image_diffusion_examples}
\end{figure}

Figure~\ref{fig:image_diffusion_process} provides a visualization of the denoising process in the reverse diffusion trajectory. Starting from pure Gaussian noise, the model progressively refines the image over 1,000 steps in our setup, gradually introducing particle-specific structures in a visually interpretable manner. To illustrate this process, we present six representative images along the trajectory. This progression highlights the ability of the generative model to recover subtle class-specific features, such as the halo effect in air bubbles or the smooth texture in silicone oil droplets, reinforcing confidence in the utility of these samples for downstream classification.

The visual quality of the generated samples from the fully trained models is illustrated in Figure~\ref{fig:image_diffusion_examples}. As can be seen, the synthesized images closely resemble real FIM images, capturing essential morphological features such as shape irregularities, texture, and transparency. Notably, the generated air bubble images preserve the circular and semi-translucent characteristics typically observed in real samples, while silicone oil droplets exhibit the distinct contour and scattering patterns that differentiate them from other SvP types. These qualitative results suggest that our diffusion models effectively learns the underlying distribution of each particle type, despite the low data regime.

\begin{table}[t]
\centering
\caption{Classification performance on the imbalanced validation set across various training configurations. ``Real" datasets contain only the images obtained through FIM , while ``Mix" datasets include both FIM and diffusion-generated samples. Results are reported for both ResNet-50 and ResNet-18 models using per-class precision (Prec.), macro-averaged precision, and area under precision-recall curve (AUPRC). For macro-averaged precision and AUPRC, we highlight the best-performing model for the respective dataset split (Real vs Mixed) with bold font.}
\scriptsize
\label{tbl:classification_results}
\begin{tabular}{ccccc|cc}
\toprule
\multirow{4}{*}{\shortstack{Model}} & \multirow{4}{*}{\shortstack{Training\\Dataset}} & \multicolumn{4}{c}{Performance on Validation Set} \\
\cmidrule[0.75pt]{3-7}
~ & ~ & \multicolumn{4}{c}{Precision} & \multirow{2}{*}{\shortstack{AUPRC}}
% ~ & ~ & Silicone Oil & Air Bubble & Protein & Macro & \multirow{2}{*}{\shortstack{AUPRC}}
\\[5pt]  % adds vertical space similar to cmidrule
~ & ~ & Silicone Oil & Air Bubble & Protein & Macro & ~ \\
\midrule
\multirow{10}{*}{\shortstack{ResNet-18}} & Real-0 & 94.00 & 98.80 & 98.06 & 96.95 & 89.34 \\
\cmidrule[0.75pt]{2-7}
~ & Real-1 & 91.60 & 99.00 & 99.24 & 96.61 & 92.45 \\
~ & Real-2 & 92.40 & 97.60 & 99.42 & 96.47 & 93.01 \\
~ & Real-3 & 92.00 & 99.00 & 97.86 & 96.29 & 90.32 \\
~ & Real-4 & 87.80 & 98.60 & 99.47 & 95.29 & 94.66 \\
\cmidrule[0.75pt]{2-7}
~ & Mixed-1 & 97.40 & 98.20 & 96.83 & \textbf{97.48} & \textbf{92.46} \\
~ & Mixed-2 & 95.20 & 98.60 & 97.96 & \textbf{97.25} & \textbf{93.09} \\
~ & Mixed-3 & 95.00 & 98.40 & 98.92 & \textbf{97.44} & \textbf{95.45} \\
~ & Mixed-4 & 95.00 & 99.00 & 98.61 & \textbf{97.54} & \textbf{96.96} \\
\midrule
\multirow{10}{*}{\shortstack{ResNet-50}}  & Real-0 & 96.35 & 95.00 & 97.80 & 96.38 & 88.48 \\
\cmidrule[0.75pt]{2-7}
~ & Real-1 & 99.18 & 90.00 & 98.60 & 95.93 & \textbf{94.85} \\
~ & Real-2 & 99.42 & 89.60 & 97.00 & 95.34 & \textbf{95.20} \\
~ & Real-3 & 98.22 & 93.80 & 98.40 & 96.81 & 92.53 \\
~ & Real-4 & 98.40 & 90.80 & 98.40 & 96.17 & 94.77 \\
\cmidrule[0.75pt]{2-7}
~ & Mixed-1 & 95.60 & 99.00 & 96.91 & \textbf{97.17} & 92.21 \\
~ & Mixed-2 & 94.80 & 99.00 & 98.32 & \textbf{97.37} & 95.15 \\
~ & Mixed-3 & 96.60 & 97.40 & 98.89 & \textbf{97.63} & \textbf{96.93} \\
~ & Mixed-4 & 95.20 & 98.20 & 99.39 & \textbf{97.60} & \textbf{97.51} \\
\bottomrule
\end{tabular}
\end{table}

\subsection{Classification Model Evaluation}

In order to quantify the usefulness of AI-generated SvP images, we conduct a large-scale classification experiment using two deep learning models: ResNet-18 and ResNet-50. As described in Section~\ref{sec:dataset}, we evaluated the classification performance of models on a heavily imbalanced validation set, using multiple training configurations with and without diffusion-generated images. Two architectures, ResNet-18 and ResNet-50, were trained on progressively larger real-only datasets (Real-n) and their augmented counterparts (Mixed-n), allowing us to assess the impact of generated data across different model capacities and data regimes. The results of these experiments, along with evaluation metrics including per-class precision, macro-averaged precision, and AUPRC, are provided in Table~\ref{tbl:classification_results}.

Across both architectures, the addition of diffusion-generated images led to consistent improvements in predictive performance. For instance, in the ResNet-50 experiments, Mixed-3 outperformed its real-only counterpart Real-3 by a margin of 0.82 in macro precision (97.63\% vs. 96.81\%) and 4.4 points in AUPRC (96.93 vs. 92.53). Similar trends were observed across all Mixed-$n$ variants, with the largest gain observed in Mixed-4, which achieved the highest AUPRC score of 97.51.

For the ResNet-50 architecture, the most notable improvements were observed in the Mixed-3 and Mixed-4 datasets. Compared to their real-only counterparts, these models achieved macro-averaged precision scores exceeding 97.6\%, with AUPRC values reaching up to 97.51. These results indicate that adding synthetic images effectively bridges the gap in predictive performance caused by class imbalance. Interestingly, gains in predictive performance plateaued between Mix-3 and Mixed-4, suggesting that beyond a certain point, additional synthetic data yields diminishing returns.

ResNet-18, despite its shallower architecture, showed similar trends. The best Mixed-n configurations for ResNet-18 achieved macro-precision scores above 97.4 and AUPRC values close to 97.0, closely trailing the deeper ResNet-50 model, suggesting that the benefits of data augmentation extend across different model capacities.

\subsection{Misclassified Protein Particles}

\begin{figure}[t!]
\centering
\includegraphics[width=0.95\textwidth]{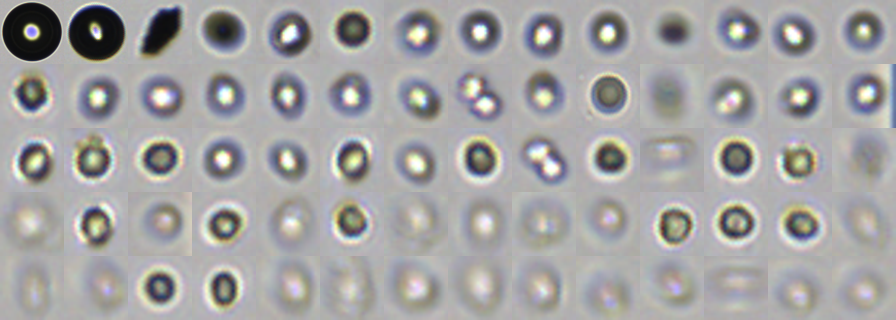}
\caption{SvP images labeled as ``protein particles" that are consistently misclassified by our classification models.}
\label{fig:misclassification_examples}
\end{figure}

In Figure~\ref{fig:misclassification_examples}, we present several SvP images labeled as 'protein particles' that are consistently misclassified by our models as silicone oil droplets or air bubbles. Upon initial inspection, many of these particles appear spherical or transparent which are visual traits more commonly associated with silicone oil or air bubbles than protein aggregates. After expert review, we conclude that several of these images are indeed more likely to be silicone oil droplets or air bubbles, indicating possible inaccuracies in the original ground truth labels. In other cases, particularly for the blurry or amorphous particles, the true identity remains uncertain, as such morphology could correspond to protein particles. These observations demonstrate that our model is highly sensitive to subtle morphological differences and, in some cases, may even outperform the original annotations by correctly reclassifying mislabeled examples.

\section{Discussion}

Our findings demonstrate that diffusion-based generative models provide an effective and scalable solution to the data imbalance problem commonly encountered in SvP classification. By generating high-fidelity synthetic images of underrepresented classes, this approach enables the construction of balanced training datasets without the need for labor-intensive manual annotation. As a result, it becomes possible to fully leverage the strengths of modern multi-class classifiers, which typically outperform one-class classifiers when sufficient class diversity is present in the training data.

An additional advantage of our method is its adaptability. While this study focused on silicone oil droplets and air bubbles in flow imaging microscopy, the same framework can be applied to other particle types, imaging modalities, or industrial quality control tasks where minority classes are difficult or expensive to annotate. Diffusion models trained on a small set of real samples can effectively synthesize realistic images that capture critical morphological variations, offering a versatile tool for balancing datasets in domains with scarce or sensitive data.

Finally, our work highlights the broader role of generative AI in pharmaceutical manufacturing and quality assurance. By reducing dependence on manual annotation, facilitating reproducible training datasets, and improving classification robustness, diffusion-based generative models have the potential to streamline quality control pipelines and support regulatory compliance in a scalable and data-driven manner. Beyond addressing the immediate challenge of data imbalance in SvP classification, our approach lays the groundwork for the broader adoption of generative AI in biomedical imaging, enabling robust and generalizable machine learning solutions in safety-critical domains.

To promote reproducible research and support future studies, we publicly release our generated datasets, trained diffusion models, and multi-class classifiers at \href{https://github.com/utkuozbulak/svp-generative-ai}{github.com/utkuozbulak/svp-generative-ai}.

\section{Conclusions}

In this study, we developed a generative AI approach using diffusion models to address data imbalance in SvP classification. Using synthesized images of underrepresented particle types, we augmented existing SvP datasets. Our experimental results demonstrate that incorporating these synthetic images into the training data leads to more balanced datasets and improved predictive performance.

Our research highlights the potential of diffusion models to enhance pharmaceutical quality control by reducing reliance on manual annotation and enabling more effective use of machine learning techniques. We anticipate that generative AI will play an increasingly important role in pharmaceutical quality control, offering scalable solutions for data scarcity and enabling more reliable, data-driven decision-making.

\section*{Acknowledgments}

M.C. is a doctoral fellow from the Research Foundation-Flanders (FWO-V) (grant number 1SH1S24N-7021).

\bibliography{main}
\end{document}